\documentclass[final]{cvpr}
\usepackage{url}
\usepackage{graphicx}
\usepackage{tabularx}
\usepackage{comment}
\usepackage{makecell}
\usepackage{pifont}
\usepackage{multirow}
\usepackage{subfigure}
\usepackage[misc]{ifsym}
\usepackage{afterpage}
\usepackage{setspace}
\usepackage{breakcites}
\usepackage{seqsplit}

\usepackage{times}
\usepackage{epsfig}
\usepackage{graphicx}
\usepackage{amsmath}
\usepackage{amssymb}
\usepackage{wrapfig}
\usepackage{hyphenat}
\usepackage{soul,color}
\usepackage{amsopn}
\usepackage{amsmath}
\usepackage{amssymb}
\usepackage{tabularx}
\usepackage{subfigure}

\definecolor{Red}{cmyk}{0,1,1,0}
\definecolor{Green}{cmyk}{1,0,1,0}
\definecolor{Cyan}{cmyk}{1,0,0,0}
\definecolor{Purple}{cmyk}{0.45,0.86,0,0}
\definecolor{Rosolic}{cmyk}{0.00,1.00,0.50,0}
\definecolor{Blue}{cmyk}{1.00,1.00,0.00,0}
\definecolor{BlueViolet}{cmyk}{0.86,0.91,0,0.04}
\definecolor{NavyBlue}{cmyk}{0.94,0.54,0,0}

\newcommand{\myparagraph}[1]{\vspace{0.1em}\noindent\textbf{#1}}

\usepackage[pagebackref=true,breaklinks=true,colorlinks,bookmarks=false]{hyperref}

\def\cvprPaperID{2221} 
\def\confYear{CVPR 2021}

\begin{document}

\title{ChallenCap: Monocular 3D Capture of Challenging Human Performances using Multi-Modal References}

\author{Yannan He\textsuperscript{1} \;\, Anqi Pang\textsuperscript{1} \;\, Xin Chen\textsuperscript{1} \;\, Han Liang\textsuperscript{1} \;\, Minye Wu\textsuperscript{1}\;\, Yuexin Ma\textsuperscript{1,2} \;\, Lan Xu\textsuperscript{1,2}} 

\makeatletter
\let\@oldmaketitle\@maketitle
\renewcommand{\@maketitle}{
	\@oldmaketitle
	\centering
	\vspace{-8mm}
	{\large \textsuperscript{1}ShanghaiTech University}\\
	{\large	\textsuperscript{2}Shanghai 
Engineering Research Center of Intelligent Vision and Imaging}
	\vspace{8mm}
}
\makeatother

\maketitle

\pagestyle{empty}
\thispagestyle{empty}

\begin{abstract}
Capturing challenging human motions is critical for numerous applications, but it suffers from complex motion patterns and severe self-occlusion under the monocular setting. 
In this paper, we propose ChallenCap --- a template-based approach to capture challenging 3D human motions using a single RGB camera in a novel learning-and-optimization framework, with the aid of multi-modal references.
We propose a hybrid motion inference stage with a generation network, which utilizes a temporal encoder-decoder to extract the motion details from the pair-wise sparse-view reference, as well as a motion discriminator to utilize the unpaired marker-based references to extract specific challenging motion characteristics in a data-driven manner. 
%
We further adopt a robust motion optimization stage to increase the tracking accuracy, by jointly utilizing the learned motion details from the supervised multi-modal references as well as the reliable motion hints from the input image reference.
Extensive experiments on our new challenging motion dataset demonstrate the effectiveness and robustness of our approach to capture challenging human motions.

\end{abstract}


\section{Introduction}
The past ten years have witnessed a rapid development of markerless human motion capture~\cite{Davison2001,hasler2009markerless,StollHGST2011, wang2017outdoor}, which benefits various applications such as immersive VR/AR experience, sports analysis and interactive entertainment.


\begin{figure}[tbp] 
	\centering 
	\includegraphics[width=0.9\linewidth]{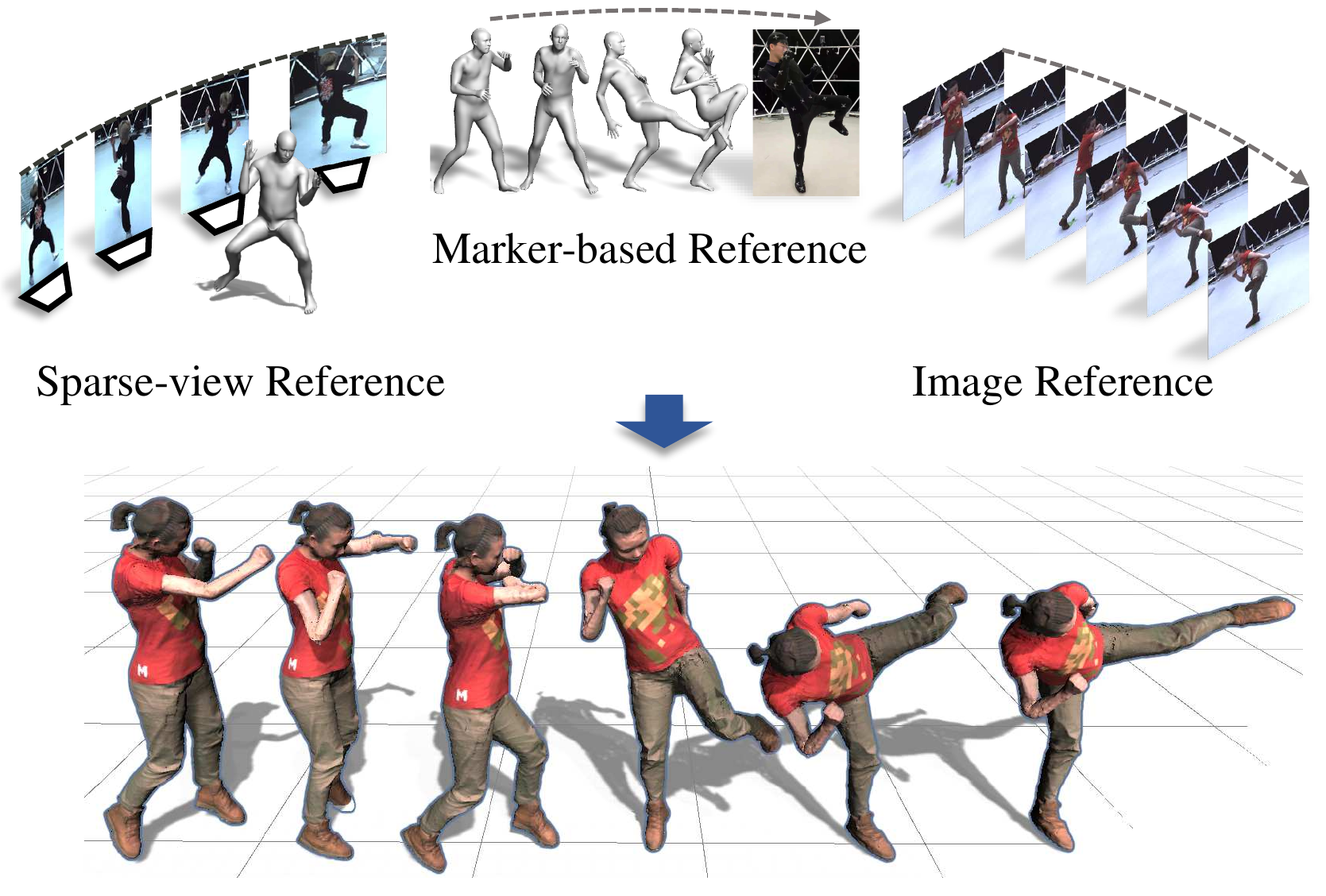} 
	\caption{Our ChallenCap approach achieves robust 3D capture of challenging human motions from a single RGB video, with the aid of multi-modal references.} 
	\label{fig:fig_1_teaser} 
	\vspace{-8pt} 
\end{figure}

Multi-view solutions~\cite{StollHGST2011,liu2013markerless,joo2015panoptic,collet2015high,TotalCapture, suo2021neuralhumanfvv} achieve high-fidelity results but rely on expensive studio setup which are difficult to be deployed for daily usage.
Recent learning-based techniques enables robust human attribute prediction from monocular RGB video~\cite{hmrKanazawa17,VIBE_CVPR2020,TEX2SHAPE_2019ICCV,DeepHuman_2019ICCV,PIFuHD}.
The state-of-the-art monocular human motion capture approaches~\cite{LiveCap2019tog,MonoPerfCap,EventCap_CVPR2020} leverage learnable pose detections~\cite{OpenPose,Mehta2017} and template fitting to achieve space-time coherent results.
However, these approaches fail to capture the specific challenging motions such as yoga or rolling on the floor, which suffer from extreme poses, complex motion patterns and severe self-occlusion under the monocular setting. 

Capturing such challenging human motions is essential for many applications such as training and evaluation for gymnastics, sports and dancing. 
Currently, optical marker-based solutions like Vicon~\cite{VICON} are widely adopted to capture such challenging professional motions.
However, directly utilizing such marker-based reference into markerless capture is inapplicable since the actor needs to re-perform the challenging motion which is temporally unsynchronized to the maker-based capture. 
Some data-driven human pose estimation approaches~\cite{HMR18,VIBE_CVPR2020} utilize the unpaired reference in an adversarial manner, but they only extract general motion prior from existing motion capture datasets~\cite{Ionescu14a,AMASS_ICCV2019}, which fails to recover the characteristics of specific challenging motion.
The recent work~\cite{DeepCap_CVPR2020} inspires to utilize the markerless multi-view reference in a data-driven manner to provide more robust 3D prior for monocular capture.
However, this method is weakly supervised on the input images instead of the motion itself, leading to dedicated per-performer training. 
%
Moreover, researchers pay less attention to combine various references from both marker-based systems and sparse multi-view systems for monocular challenging motion capture.

In this paper, we tackle the above challenges and present \textit{ChallenCap} -- a template-based monocular 3D capture approach for challenging human motions from a single RGB video, which outperforms existing state-of-the-art approaches significantly (See Fig.~\ref{fig:fig_1_teaser} for an overview).
Our novel pipeline proves the effectiveness of embracing multi-modal references from both temporally unsynchronized marker-based system and light-weight markerless multi-view system in a data-driven manner, which enables robust human motion capture under challenging scenarios with extreme poses and complex motion patterns, whilst still maintaining a monocular setup.


More specifically, we introduce a novel learning-and-optimization framework, which consists of a hybrid motion inference stage and a robust motion optimization stage. 
Our hybrid motion inference utilizes both the marker-based reference which encodes the accurate spatial motion characteristics but sacrifices the temporal consistency, as well as the sparse multi-view image reference which provides pair-wise 3D motion priors but fails to capture extreme poses.
To this end, we first obtain the initial noisy skeletal motion map from the input monocular video.
Then, a novel generation network, HybridNet, is proposed to boost the initial motion map, which utilizes a temporal encoder-decoder to extract local and global motion details from the sparse-view reference, as well as a motion discriminator to utilize the unpaired marker-based reference.
Besides the data-driven 3D motion characteristics from the previous stage, the input RGB video also encodes 
reliable motion hints for those non-extreme poses, especially for the non-occluded regions.
Thus, a robust motion optimization is further proposed to refine the skeletal motions and improve the tracking accuracy and overlay performance, which jointly utilizes the learned 3D prior from the supervised multi-modal references as well as the reliable 2D and silhouette information from the input image reference.
To summarize, our main contributions include: 
\begin{itemize} 
\setlength\itemsep{0em}
	\item We propose a monocular 3D capture approach for challenging human motions, which utilizes multi-modal reference in a novel learning-and-optimization framework, achieving significant superiority to state-of-the-arts. 
	
	\item We propose a novel hybrid motion inference module to learn the challenging motion characteristics from the supervised references modalities, as well as a robust motion optimization module for accurate tracking.
	
	\item We introduce and make available a new challenging human motion dataset with both unsynchronized marker-based and light-weight multi-image references, covering 60 kinds of challenging motions and 20 performers with 120k corresponding images.
\end{itemize}

\begin{figure*}[t]
	\centering
	\includegraphics[width=\linewidth]{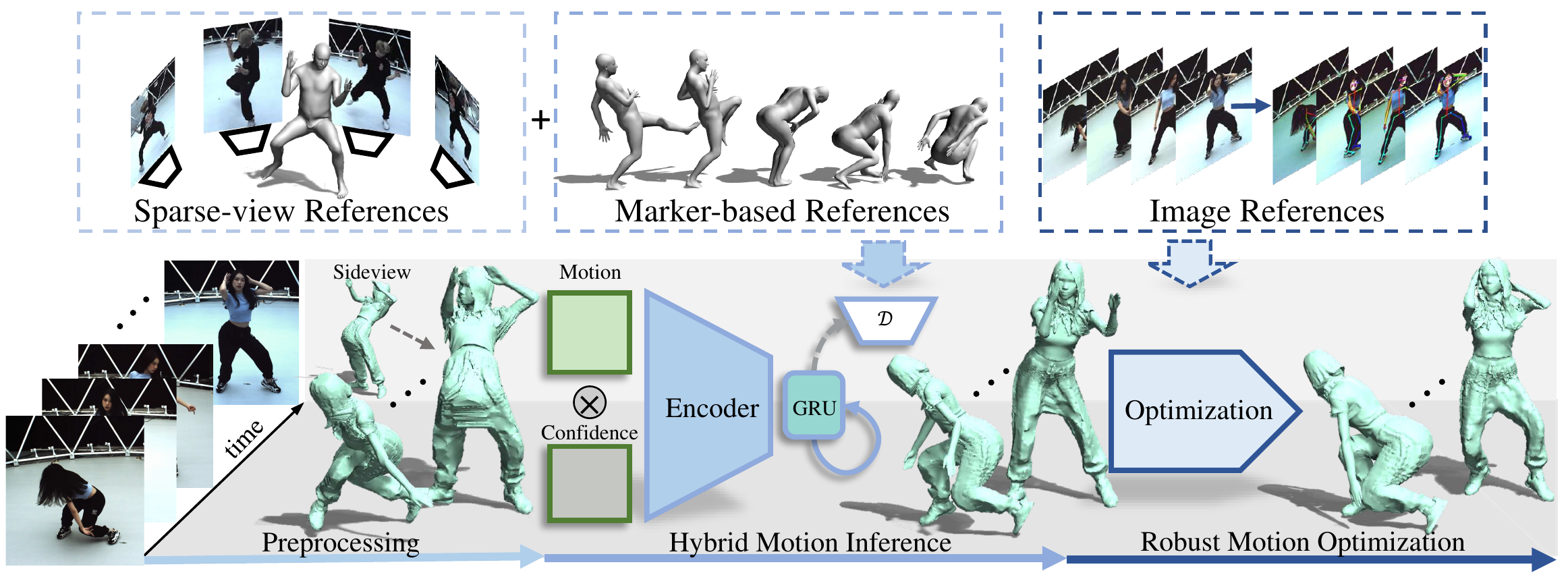}
	\caption{The pipeline of ChallenCap with multi-modal references. Assuming the video input from monocular camera, our approach consists of a hybrid motion inference stage (Sec.\ref{sec:mot_inference}) and a robust motion optimization stage (Sec.\ref{sec:mot_optimization}) to capture 3D challenging motions. $\mathcal{D}$ represents the discriminator. }  
	\label{fig:overview}
\end{figure*}

\section{Related Work}
As an alternative to the widely used marker-based solutions~\cite{VICON,Xsens,Vlasic2007}, markerless motion capture~\cite{BreglM1998,AguiaSTAST2008,TheobASST2010} technologies alleviate the need for body-worn markers and have been widely investigated.
In the following, we focus on the field of marker-less 3D human motion capture.

\myparagraph{Parametric Model-based Capture.}
Many general human parametric models~\cite{SCAPE2005,SMPL2015,SMPLX2019,STAR_ECCV2020} learned from thousands of high-quality 3D scans have been proposed in the last decades, which factorize human deformation into pose and shape components. 
Deep learning is widely used to obtain skeletal pose and human shape prior through model fitting~\cite{TAM_3DV2017,Lassner17,keepitSMPL,Kolotouros_2019_CVPR} or directly regressing the model parameters from the input~\cite{HMR18,Kanazawa_2019CVPR,VIBE_CVPR2020,zanfir2020neural}.
Besides, various approaches~\cite{detailHuman_2019CVPR,DetailDepth_2019ICCV,People3D_2019ICCV,TEX2SHAPE_2019ICCV,DeepHuman_2019ICCV} propose to predict detailed human geometry by utilizing parametric human model as a basic estimation.
Beyond human shape and pose, recent approaches further include facial and hand models~\cite{Xiang_2019_CVPR,TotalCapture,SMPLX2019,MonoExpressive_ECCV2020} for expressive reconstruction or leverage garment and clothes modeling on top of parametric human model~\cite{ClothCap,bhatnagar2019mgn,TailorNet_CVPR2020,LearnToDress_CVPR2020}.
But these methods are still limited to the parametric model and cannot provide space-time coherent results for loose clothes. 
Instead, our method is based on person-specific templates and focuses on capturing space-time coherent challenging human motions using multi-modal references.

\myparagraph{Free-form Volumetric Capture.}
Free-form volumetric capture approaches with real-time performance have been proposed by combining the volumetric fusion \cite{Curless1996} and the nonrigid tracking~\cite{sumner2007embedded,li2009robust,zollhofer2014real,guo2015robust} using depth sensors. 
The high-end solutions~\cite{dou-siggraph2016,motion2fusion,UnstructureLan,TheRelightables} rely on multi-view studios which are difficult to be deployed.
The most handy monocular approaches for general non-rigid scenes~\cite{Newcombe2015,innmann2016volume,guo2017real,KillingFusion2017cvpr,slavcheva2018cvpr,slavcheva2020pami,FlyFusion} 
can only capture small, controlled, and slow motions.
Researchers further utilize parametric model~\cite{BodyFusion,DoubleFusion,UnstructureLan,robustfusion} or extra body-worn sensors~\cite{HybridFusion} into the fusion pipeline to increase the tracking robustness.
However, these fusion approaches rely on depth cameras which are not as cheap and ubiquitous as color cameras.
Recently, the learning-based techniques enable free-form human reconstruction from monocular RGB input with various representations, such as volume~\cite{DeepHuman_2019ICCV}, silhouette~\cite{SiCloPe_CVPR2019} or implicit representation~\cite{PIFU_2019ICCV,PIFuHD,MonoPort,Chibane_2020CVPR,Implicit_ECCV2020}.
However, such data-driven approaches do not recover temporal-coherent reconstruction, especially under the challenging motion setting. 
In contrast, our template-based approach can explicitly obtain the per-vertex correspondences over time even for challenging motions.

\myparagraph{Template-based Capture.}
A good compromising settlement between from-form capture and parametric Modal-based Capture is to utilize a specific human template mesh as prior.
Early solutions~\cite{Gall2010,StollHGST2011,liu2013markerless,Robertini:2016,Pavlakos17,Simon17,FlyCap} require multi-view capture to produce high quality skeletal and surface motions but synchronizing and calibrating multi-camera systems is still cumbersome. 
Recent work only relies on a single-view setup~\cite{MonoPerfCap,LiveCap2019tog,EventCap_CVPR2020} achieve space-time coherent capture and even achieves real-time performance~\cite{LiveCap2019tog}.
However, these approaches fail to capture the challenging motions such as yoga or rolling on the floor, which suffer from extreme poses and severe self-occlusion under the monocular setting. 
The recent work~\cite{DeepCap_CVPR2020} utilizes weekly supervision on multi-view images directly so as to improve the 3D tracking accuracy during test time.
However, their training strategy leads to dedicated per-performer training.
Similarly, our approach also employs a person-specific template mesh.
Differently, we adopt a specific learning-and-optimization framework for challenging human motion capture.
Our learning module is supervised on the motion itself instead of the input images of specific performers for improving the generation performance to various performers and challenging motions. 

\begin{figure*}[t]
	\centering
	\includegraphics[width=\linewidth]{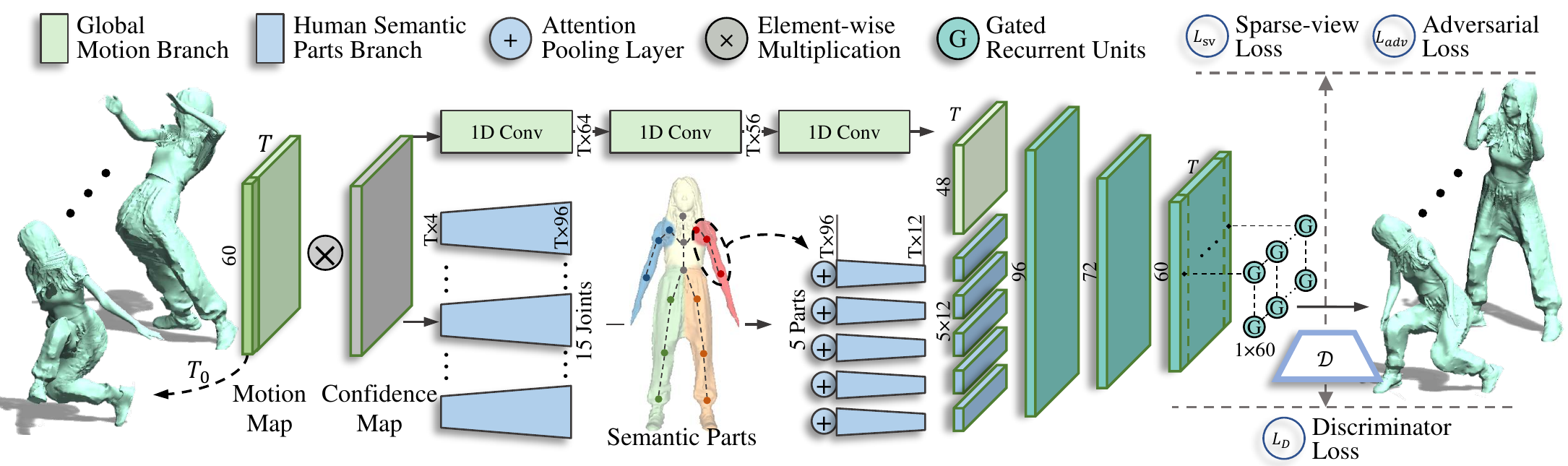}
	\caption{Illustration of our hybrid motion network,  HybridNet, which encodes the global and local temporal motion information with the losses on both the generator and the discriminator. Note that the attention pooling operation is performed by applying element-wise addition (blue branches) on the features of the adjacent body joints. The features after attention pooling are concatenated together with the global feature as input to the fully connected layers.}
	\label{fig:fig_3_network}
\end{figure*}


\section{Overview}
Our goal is to capture challenging 3D human motions from a single RGB video, which suffers from extreme poses, complex motion patterns and severe self-occlusion.
Fig.~\ref{fig:overview} provides an overview of ChallenCap, which relies on a template mesh of the actor and makes full usage of multi-modal references in a learning-and-optimization framework.
Our method consists of a hybrid motion inference module to learn the challenging motion characteristics from the supervised references modalities, and a robust motion optimization module to further extract the reliable motion hints in the input images for more accurate tracking.

\myparagraph{Template and Motion Representation.}
We use a 3D body scanner to generate the template mesh of the actor and rig it by fitting the Skinned Multi-Person Linear Model (SMPL)\cite{SMPL2015} to the template mesh and transferring the SMPL skinning weights to our scanned mesh.
The kinematic skeleton is parameterized as $\textbf{S}=[\boldsymbol{\theta}, \textbf{R},\textbf{t}]$, including the joint angles $\boldsymbol{\theta} \in \mathbb{R}^{30}$ of the $N_J$ joints, the global rotation $ \textbf{R}\in\mathbb{R}^3$ and translation $\textbf{t} \in \mathbb{R}^{3}$ of the root.
Furthermore, let $\textbf{Q}$ denotes the quaternions representation of the skeleton.
Thus, we can formulate $\textbf{S}=\mathcal{M}(\textbf{Q},\textbf{t})$ where $\mathcal{M}$ denotes the motion transformation between various representations.


\myparagraph{Hybrid Motion Inference.}
Our novel motion inference scheme extracts the challenging motion characteristics from the supervised marker-based and sparse multi-view references in a data-driven manner.
We first obtain the initial noisy skeletal motion map from the monocular video.
Then, a novel generation network, HybridNet, is adopted to boost the initial motion map, which consists of a temporal encoder-decoder to extract local and global motion details from the sparse-view references, as well as a motion discriminator to utilize the motion characteristics from the unpaired marker-based references.
To train our HybridNet, a new dataset with rich references modalities and various challenging motions is introduced (Sec. \ref{sec:mot_inference}).


\myparagraph{Robust Motion Optimization.}
Besides the data-driven 3D motion characteristics from the previous stage, the input RGB video also encodes
reliable motion hints for those non-extreme poses, especially for the non-occluded regions.
Thus, a robust motion optimization is introduced to refine the skeletal motions so as to increase the tracking accuracy and overlay performance, which jointly utilizes the learned 3D prior from the supervised multi-modal references as well as the reliable 2D and silhouette information from the input image references (Sec. \ref{sec:mot_optimization}).



\section{Approach}
\subsection{Hybrid Motion Inference} \label{sec:mot_inference}

\myparagraph{Preprocessing.} 
Given an input monocular image sequence $I_t, t\in[1,T]$ of length $T$ and a well-scanned template model,
we first adopt the off-the-shelf template-based motion capture approach~\cite{MonoPerfCap} to obtain the initial skeletal motion $\textbf{S}_t$ and transform it into quaternions format, denoted as $\textbf{Q}^{init}_t$.
More specifically, we only adopt the 2D term from \cite{MonoPerfCap} using OpenPose~\cite{OpenPose} to obtain 2D joint detections. 
Please refer to \cite{MonoPerfCap} for more optimization details.
Note that such initial skeletal motions suffer from severe motion ambiguity since no 3D prior is utilized, as illustrated in the pre-processing stage in Fig.~\ref{fig:overview}.
After the initial optimization, we concatenate the $\textbf{Q}^{init}_t$ and the detection confidence from OpenPose~\cite{OpenPose} for all the $T$ frames into a motion map $\mathcal{Q}\in\mathbb{R}^{T\times 4N_J}$ as well as a confidence map $\mathcal{C}\in\mathbb{R}^{T\times N_J}$.


\myparagraph{HybridNet Training.}
Based on the initial noisy motion map $\mathcal{Q}$ and confidence map $\mathcal{C}$, we propose a novel generation network, HybridNet, to boost the initial capture results for challenging human motions.

As illustrated in Fig.~\ref{fig:fig_3_network}, our HybridNet learns the challenging motion characteristics by the supervision from multi-modal references.
To avoid tedious 3D pose annotation, we utilize the supervision from the optimized motions using sparse multi-view image reference.
Even though such sparse-view reference still cannot recover all the extreme poses, it provides rich pair-wise overall 3D motion prior. 
To further extract the fine challenging motion details, we utilize adversarial supervision from the marker-based reference since it only provides accurate but temporally unsynchronized motion characteristics due to re-performing. 
To this end, we utilize the well-known Generative Adversarial Network (GAN) structure in our HybridNet with the generative network $\mathcal{G}$ and the discriminator network $\mathcal{D}$.

Our generative module consists of a global-local motion encoder with a hierarchical attention pooling block and a GRU-based decoder to extract motion details from the sparse-view references, which takes the concatenated $\mathcal{Q}$ and $\mathcal{C}$ as input.
In our encoder, we design two branches to encode the global and local skeletal motion features independently.
Note that we mainly use 1D convolution layers during the encoding process to extract corresponding temporal information.
We apply three layers of 1D convolution for the global branch while splitting the input motion map into $N_J$ local quaternions for the local branch inspired by \cite{taew}.
Differently, we utilize a hierarchy attention pooling layer to connect the feature of adjacent joints and compute the latent codes from five local body regions, including the four limbs and the torso.
We concatenate the global and local feature of two branches as the final latent code, and decode them to the original quaternions domain with three linear layers in our GRU-based decoder (see Fig.~\ref{fig:fig_3_network} for detail).
Here, the loss of our generator $\mathcal{G}$ is formulated as:
\begin{align}
\mathcal{L}_{\mathcal{G}} = \mathcal{L}_{sv} + \mathcal{L}_{adv},
\end{align}
where $\mathcal{L}_{sv}$ is the sparse-view loss and $\mathcal{L}_{adv}$ is the adversarial loss.
Our sparse-view loss is formulated as:
\begin{align}
& \mathcal{L}_{sv} = \sum_{t=1}^T \left\|\hat{\textbf{Q}}_t - \textbf{Q}^{sv}_t\right\|_2^2 + \lambda_{quat}\sum_{t=1}^T\sum^{N_J}_{i=1}(\|\hat{\textbf{Q}}_t^{(i)}\| - 1)^2.
\end{align}
Here, the first term is the $L_2$ loss between the regressed output motion $\hat{\textbf{Q}}_t$ and the 3D motion prior $\textbf{Q}^{sv}_t$ from sparse-views reference.
Note that we obtain $\textbf{Q}^{sv}_t$ from the reference sparse multi-view images by directly extending the same optimization process to the multi-view setting.
The second regular term forces the network output quaternions to represent a rotation.
Besides, $N_J$ denotes the number of joints which is 15 in our case while $\lambda_{quat}$ is set to be $1\times 10^{-5}$.

Our motion discriminator $\mathcal{D}$ further utilizes the motion characteristics from the unpaired marker-based references, which maps the motion map $\mathcal{\hat{Q}}$ corrected by the generator to a value $\in [0, 1]$ to represent the probability $\mathcal{\hat{Q}}$ is a plausible human challenging motion.
Specifically, we follow the video motion capture approach VIBE~\cite{VIBE_CVPR2020} to design two losses, including the adversarial loss $\mathcal{L}_{adv}$ to backpropagate to the generator $\mathcal{G}$ and the discriminator loss $\mathcal{L}_{\mathcal{D}}$ for the discriminator $\mathcal{D}$:
\begin{align}
    &\mathcal{L}_{adv} = \mathbb{E}_{\mathcal{\hat{Q}} \sim p_{\mathcal{G}}}\left[\left(\mathcal{D}(\mathcal{\hat{Q}})-1\right)^{2}\right],\\
	&\mathcal{L}_{\mathcal{D}} = \mathbb{E}_{\mathcal{Q}^{mb} \sim p_{V}}\left[\left(\mathcal{D}(\mathcal{Q}^{mb})-1\right)^{2}\right]+\mathbb{E}_{\mathcal{\hat{Q}} \sim p_{\mathcal{G}}}\left[\left(\mathcal{D}(\mathcal{\hat{Q}})\right)^{2}\right].
\end{align}
Here, the adversarial loss $\mathcal{L}_{adv}$ is the expectation that $\mathcal{\hat{Q}}$ belongs to a plausible human challenging motion, while $p_{\mathcal{G}}$ and $p_V$ represents the corrected motion sequence the corresponding captured maker-based challenging motion sequence, respectively.
Note that $\mathcal{Q}^{mb}$ denotes the accurate but temporally unsynchronized motion map captured by the marker-based system.
Compared to VIBE~\cite{VIBE_CVPR2020} which extracts general motion prior, our scheme can recover the characteristics of more specific challenging motion.

\myparagraph{Training Details.}
We train our HybridNet for 500 epochs with Adam optimizer~\cite{kingma2014adam}, and set the dropout ratio as 0.1 for the GRU layers.
We apply Exponential Linear Unit (ELU) activation and batch normalization layer after every our 1D convolutional layer with kernel size 7, except the final output layer before the decoder.
During training, four NVidia 2080Ti GPUs are utilized.
The batch size is set to be 32, while the learning rate is set to be $1 \times 10^{-3}$ for the generator and $1 \times 10^{-2}$ for the discriminator, the decay rate is 0.1 (final 100 epochs).
To train our HybridNet, a new dataset with rich references modalities and various challenging motions and performers is further introduced and more details about our dataset are provided in Sec.~\ref{ExperimentalResults}

Our hybrid motion inference utilizes multi-modal references to extract fine motion details for challenging human motions in a data-driven manner.
At test time, our method can robustly boost the tracking accuracy of the initial noisy skeletal motions via a novel generation network.
Since our learning scheme is not directly supervised on the input images of specific performers, it's not restricted by per-performer training.
Instead, our approach focus on extracting the characteristics of challenging motions directly.


\begin{figure*}
	\centering
	\includegraphics[width=0.95\linewidth]{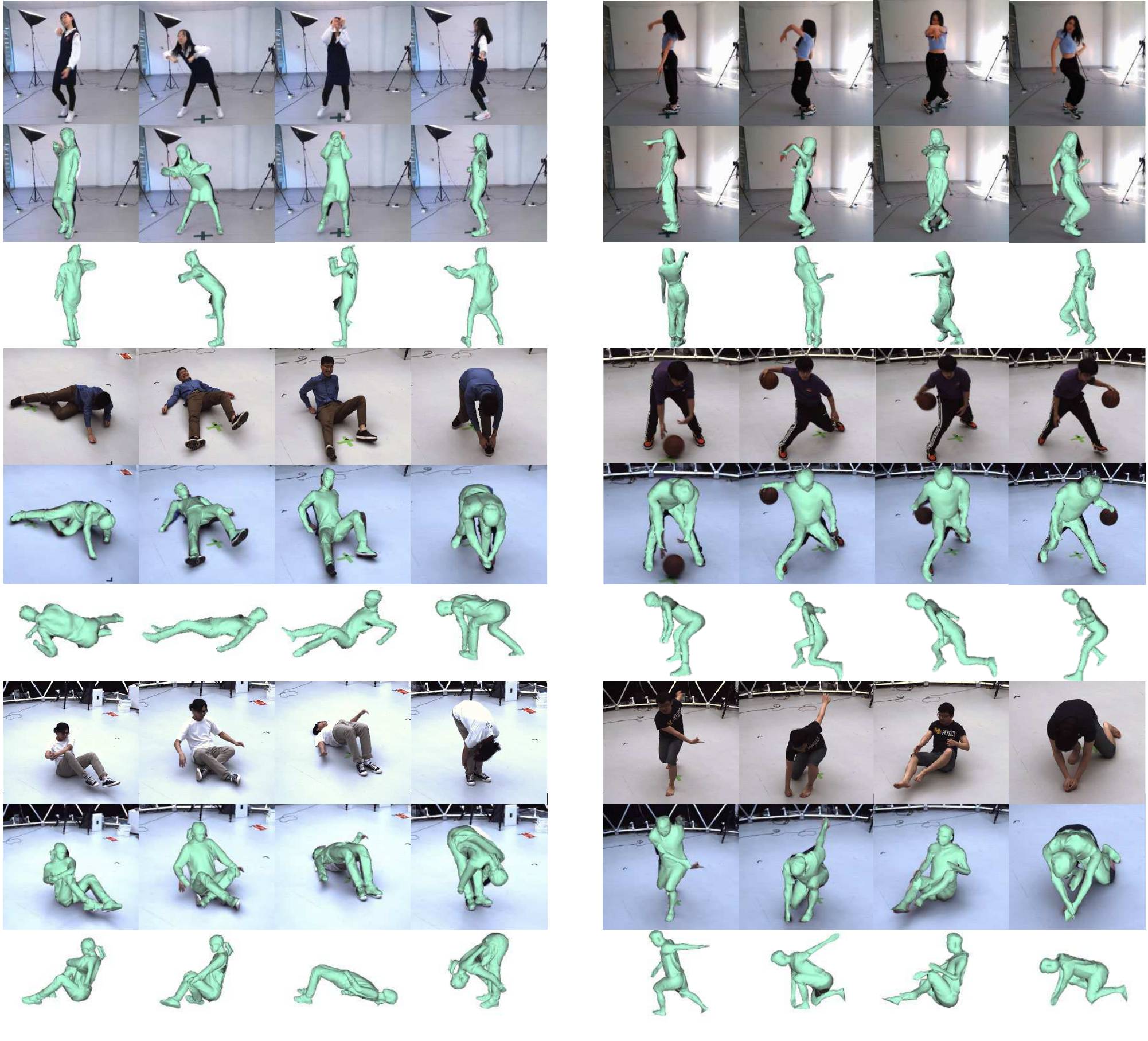}
	\caption{3D capturing results on challenging human motions. For each body motions, the top row shows the input images, the middle row shows the captured body results on camera view, and the bottom row shows the rendering result of the captured body from side view. }
	\label{fig:rs_gallery}
\end{figure*}

\subsection{Robust Motion Optimization}
\label{sec:mot_optimization}
Besides the data-driven 3D motion characteristics from the previous stage, the input RGB video also encodes reliable motion hints for those non-extreme poses, especially for the non-occluded regions. We thus introduce this robust motion optimization to refine the skeletal motions so as to increase the tracking accuracy and overlay performance, which jointly utilizes the learned 3D prior from the supervised multi-modal references as well as the reliable 2D and silhouette information from the input image references.
The optimization to refine the skeletal pose is formulated as:
\begin{align} \label{eq:opt}
\boldsymbol{E}_{\mathrm{total}}(\textbf{S}_t) = \boldsymbol{E}_{\mathrm{3D}} + \lambda_{\mathrm{2D}}\boldsymbol{E}_{\mathrm{2D}} + \lambda_{\mathrm{T}}\boldsymbol{E}_{\mathrm{T}} + \lambda_{\mathrm{S}}\boldsymbol{E}_{\mathrm{S}}.
\end{align}
Here, $\boldsymbol{E}_{\mathrm{3D}}$ makes the final motion sequence close to the output of network on occluded and invisible joints while  $\boldsymbol{E}_{\mathrm{2D}}$ adds a re-projection constraint on high-confidence 2D keypoints detected. $\boldsymbol{E}_{\mathrm{T}}$ enforces the final motion to be temporally smooth, while the $\boldsymbol{E}_{\mathrm{S}}$ enforces alignment of the projected 3D model boundary with the detected silhouette.
%
Specifically, the 3D term $\boldsymbol{E}_{\mathrm{3D}}$ is as following:
\begin{align}
\boldsymbol{E}_{\mathrm{3D}} = \sum^T_{t=1}\|\textbf{S}_t - \mathcal{M}(\hat{\textbf{Q}}_t,\textbf{t}_t)  \|^2_2,
\end{align}
where $\hat{\textbf{Q}}_t$ is the regressed quaternions motion from our previous stage; $\mathcal{M}$ is the mapping from quaternions to skeletal poses; $\textbf{t}_t$ is the global translation of $\textbf{S}_t$.
Note that the joint angles $\boldsymbol{\theta}_t$ of $\textbf{S}_t$ locate in the pre-defined range $[\boldsymbol{\theta}_{min},  \boldsymbol{\theta}_{max}]$ of physically plausible joint angles to prevent unnatural poses.
We then propose the projected 2D term as:
\begin{align}
\boldsymbol{E}_{\mathrm{2D}} = \frac{1}{T}\sum^T_{t=1}\frac{1}{|C_t|}\sum_{i\in C_t
}\|\Pi(J_i(\textbf{S}_t))-\mathbf{p}_{t}^{(i)}\|^2_2,
\end{align}
where $C_t = \{ i \ |\  c_t^{(i)} \geq thred \}$ is the set of indexes of high-confidence keypoints on the image $I_t$; $ c_t^{(i)}$ is the confidence value of the $i^{th}$ keypoint $\mathbf{p}_{t}^{(i)}$; $thred$ is 0.8 in our implementation. 
The projection function $\Pi$ maps 3D joint positions to 2D coordinates while $J_i$ computes the 3D position of the $i^{th}$ joint. 
Then, the temporal term $\boldsymbol{E}_{\mathrm{T}}$ is formulated as:
\begin{align}
\boldsymbol{E}_{\mathrm{T}} = \sum^{T-1}_{t=1}\|\mathcal{M}(\hat{\textbf{Q}}_t,\textbf{t}_t) - \mathcal{M}(\hat{\textbf{Q}}_{t+1},\textbf{t}_{t+1})\|^2_2,
\end{align}
where $\hat{\textbf{Q}}_t$ and $\hat{\textbf{Q}}_{t+1}$ are two adjacent regressed quaternions motion from our hybrid motion inference module. 
We utilize temporal smoothing to enable globally consistent capture in 3D space.
Moreover, we follow \cite{MonoPerfCap} to formulate the silhouette term $\boldsymbol{E}_{\mathrm{S}}$ and please refer to \cite{MonoPerfCap} for more detail.

The constrained optimization problem to minimize the Eqn.~\ref{eq:opt} is solved using the Levenberg-Marquardt (LM) algorithm of  
\textit{ceres}~\cite{ceresSolver}.
In all experiments, we use the following empirically determined parameters: $\lambda_{\mathrm{2D}} = 1.0$, $\lambda_{\mathrm{T}} = 20.0$ and $\lambda_{\mathrm{S}} = 0.3$.
Note that the initial $\textbf{t}_t$ for the mapping from quaternions to skeletal poses is obtained through the pre-processing stage in Sec.~\ref{sec:mot_inference}.
To enable more robust optimization, we first optimize the global translation $\textbf{t}_t$ for all the frames and then optimize the $\textbf{S}_t$.
Such a flip-flop optimization strategy improves the overlay performance and tracking accuracy, which jointly utilizes the 3D challenging motion characteristics from the supervised multi-modal references as well as the reliable 2D and silhouette information from the input image references.

\section{Experimental Results}
\label{ExperimentalResults}
In this section, we introduce our new dataset and evaluate our ChallenCap in a variety of challenging scenarios. 

\begin{figure}[htb]
	\centering
	\includegraphics[width=\linewidth]{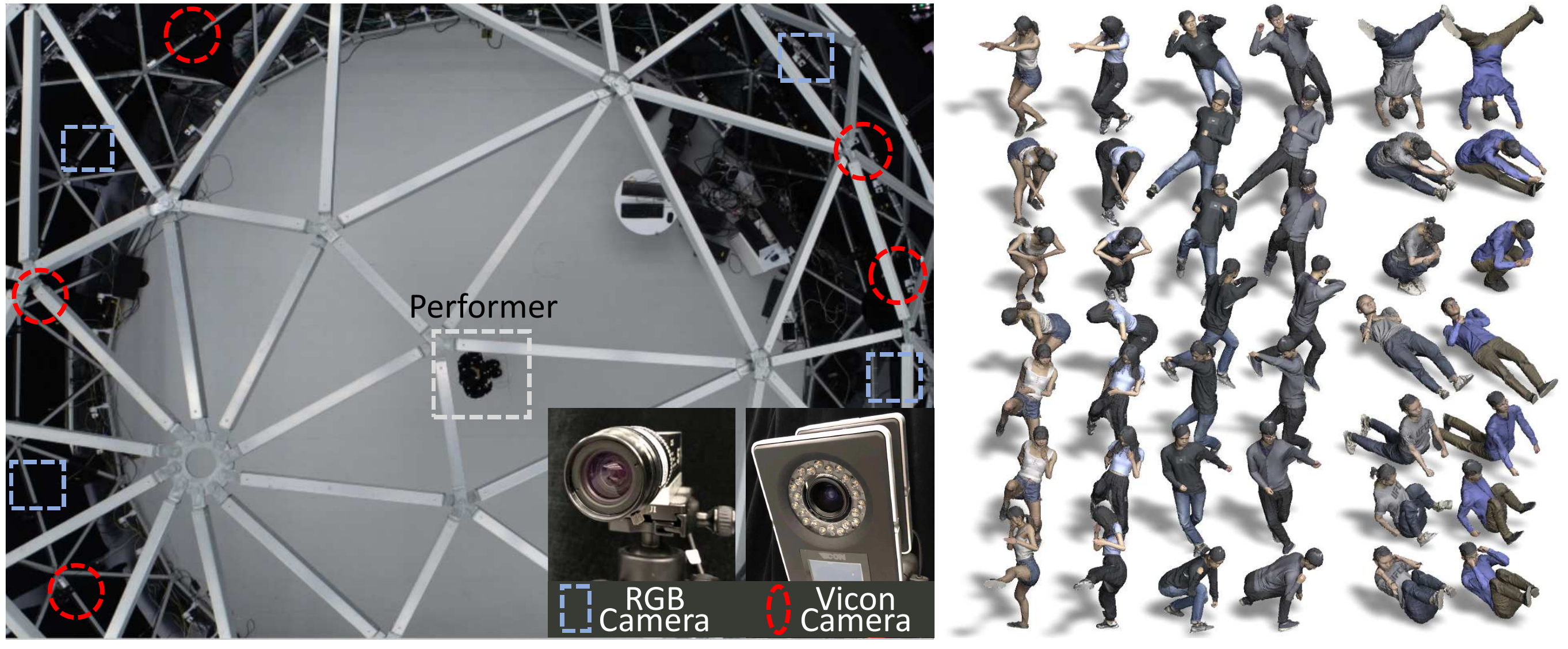}
	\caption{Illustration of our capturing system and examples of our dataset. The left shows our capturing system, including four RGB cameras (blue) for sparse-view image sequences and Vicon cameras (red) for marker-based motion capture (partially annotated). The right shows the animated meshes from the rigged character-wise template models with various challenging motions.}
	\label{fig:dataset}
\end{figure}

\myparagraph{ChallenCap Dataset.}
There are existing datasets for 3D pose estimation and human performance capture, such as the Human3.6M\cite{ionescu2013human3} dataset that contains 3D human poses in daily activities, but it lacks challenging motions and template human meshes. The AMASS \cite{AMASS_ICCV2019} dataset provides a variety of human motions using marker-based approach, but the corresponding RGB videos are not provided. 
To evaluate our method, we propose a new challenging human motion dataset containing 20 different characters with a wide range of challenging motions such as dancing, boxing, gymnastic, exercise, basketball, yoga, rolling, leap, etc(see Fig.\ref{fig:dataset}). We adopt a 12-view marker-based Vicon motion capture system to capture challenging human motions. Each challenging motion consists of data from two modalities: synchronized sparse-view RGB video sequences and marker-based reference motion captured in an unsynchronized manner. 

\begin{figure}[t]
		\centering
		\includegraphics[width=0.95\linewidth]{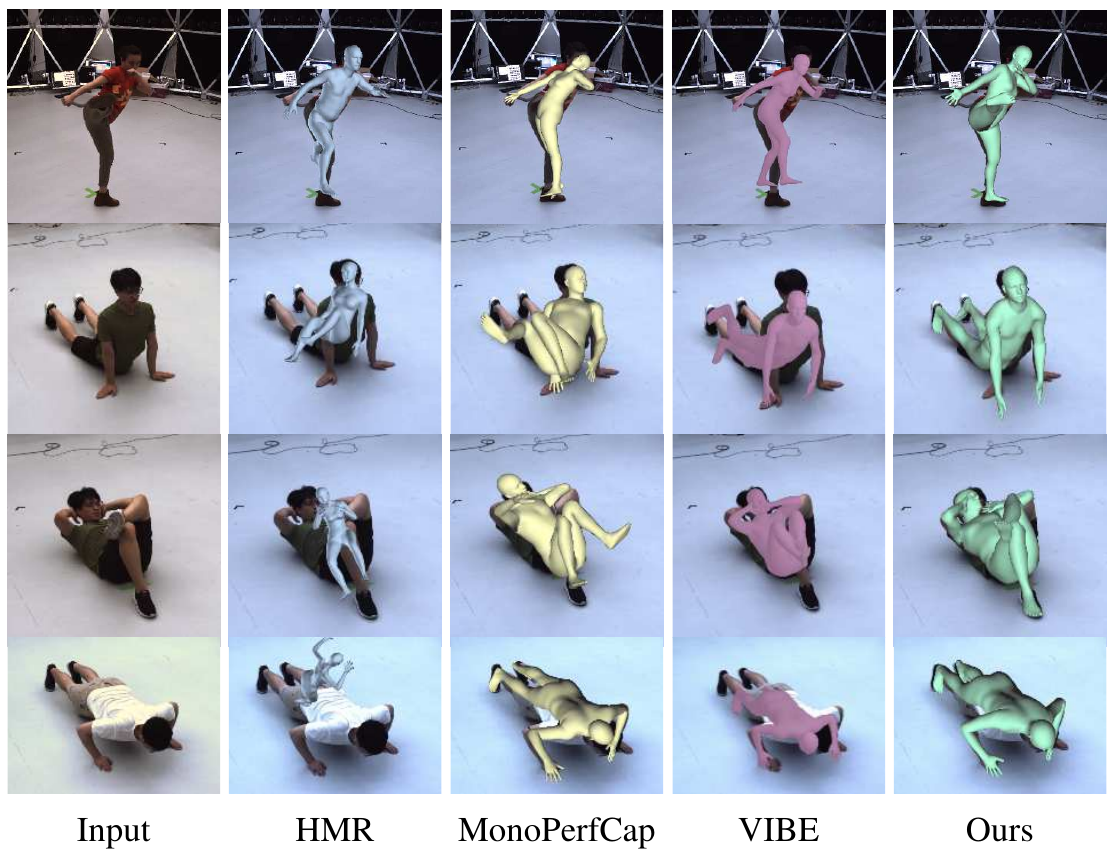}
		\caption{Qualitative comparison. Our results overlay better with the input video frames than the results of other methods.
		}
		\label{fig:comp_methods}
\end{figure}

\begin{table}[t]
	\begin{center}
		\centering
		\caption{Quantitative comparision of several methods in terms of tracking accuracy and template mesh overlay.}
		\label{tab:Comparison}
		\resizebox{0.48\textwidth}{!}{
		\begin{tabular}{l|cccc}
			\hline
			Method      & MPJPE ($mm$)$\downarrow$ & PCK0.5($\%$)$\uparrow$ & PCK0.3$\uparrow$ & mIoU($\%$)$\uparrow$ \\
			\hline
			HMR~\cite{HMR18}         & 154.3  & 77.2   & 68.9   & 57.0 \\
			VIBE~\cite{VIBE_CVPR2020}        & 116.7 & 83.7   & 71.8   & 73.7 \\
			MonoPerfCap~\cite{MonoPerfCap} & 134.7 & 77.4   & 65.6   & 65.5 \\
			Ours        & \textbf{ 52.6}  & \textbf{96.6}   & \textbf{87.4}   & \textbf{83.6} \\
			\hline
		\end{tabular}
		}
	\end{center}
\end{table}

\subsection{Comparison}

Our method enables more accurate motion capture for challenging human motions. For further comparison, we do experiments to demonstrate its effectiveness. We compare the proposed ChallenCap method with several monocular 3D human motion capture methods. 
Specifically, we apply MonoPerfCap~\cite{MonoPerfCap} which is based on optimization. We also apply HMR~\cite{HMR18} and VIBE~\cite{VIBE_CVPR2020} where the latter also relies on an adversarial learning framework. For fair comparisons, we fine-tune HMR and VIBE with part of manually annotated data from our dataset.
As shown in Fig.\ref{fig:comp_methods}, our method outperforms other methods in motion capture quality. Benefiting from multi-modal references, our method performs better on the overall scale and also gets better overlays of the captured body.

\begin{figure}
		\centering
		\includegraphics[width=0.95\linewidth]{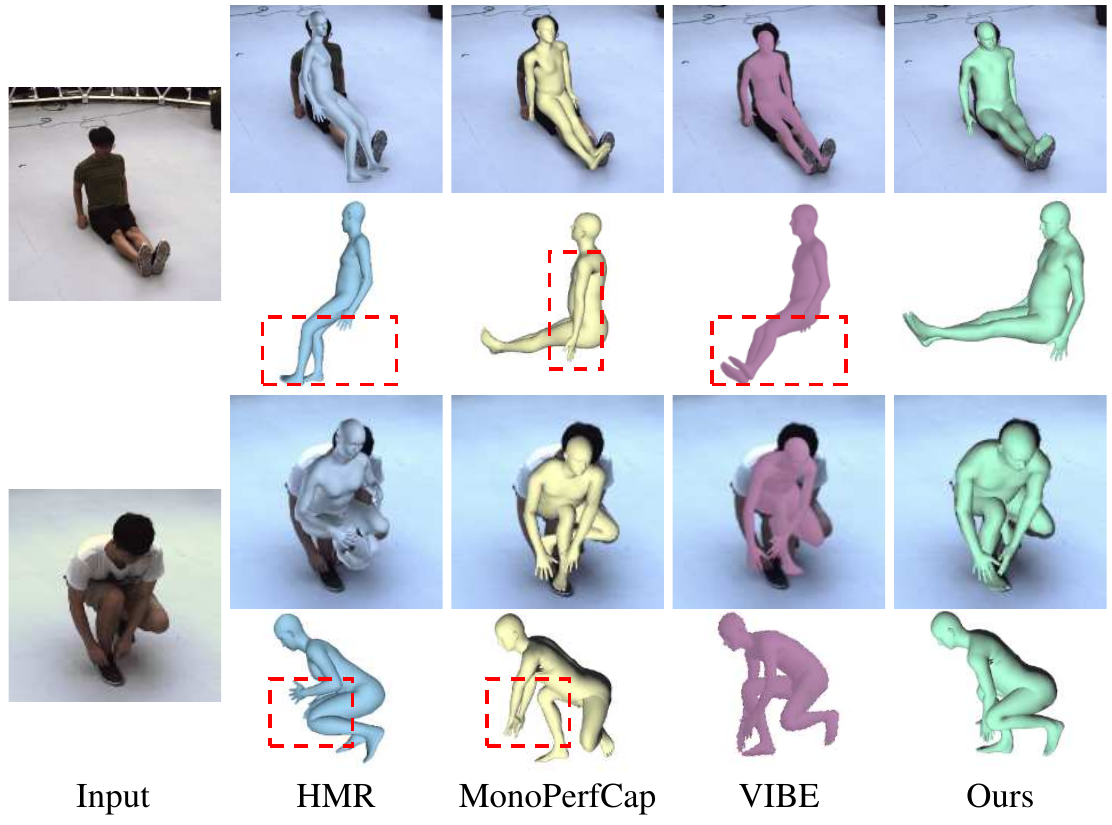}
		\caption{Qualitative comparison with side views. Our method maintains projective consistency in the side views while other methods have misalignment errors.
		}
		\label{fig:comp_sideview}
\end{figure}

As illustrated in Fig.\ref{fig:comp_sideview}, we perform a qualitative comparison with other methods on the main camera view and a corresponding side view. The figure shows that the side view results of other methods wrongly estimated the global position of arms or legs. This is mainly because our HybridNet promotes capture results for challenging human motions in the world frame.

\begin{figure}[t]
	\centering
	\includegraphics[width=0.95\linewidth]{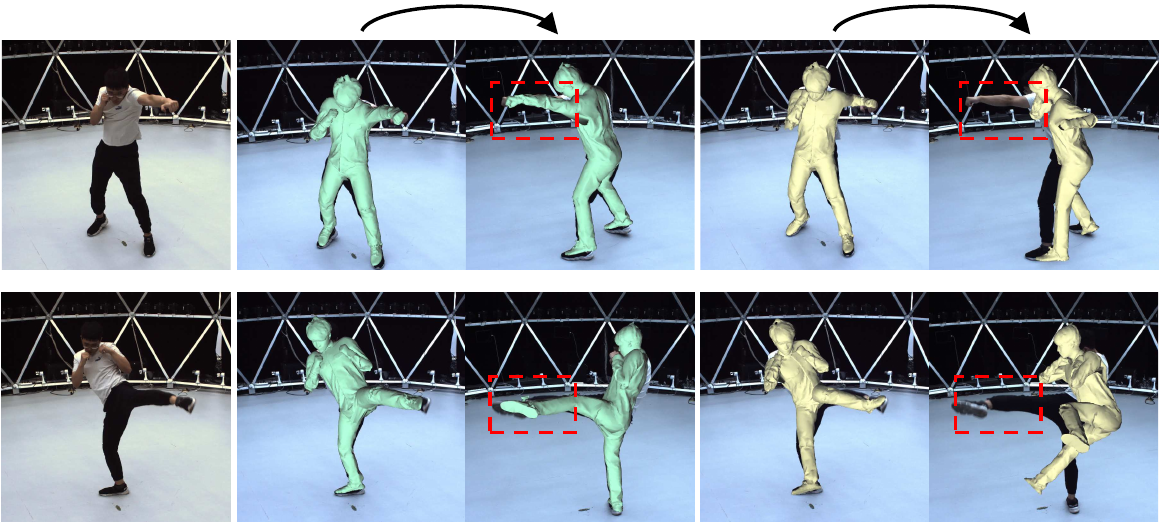}
	\caption{Qualitative comparison between ChallenCap (green) and MonoPerfCap (yellow) with reference-view verification. As marked in the figure, MonoPerfCap misses limbs in the reference view.}
	\label{fig:comp_multiview}
\end{figure}

Fig.\ref{fig:comp_multiview} shows the reference view verification results. We capture challenging motions in the main view and verify the result in a reference view with a rendered mesh. The figure shows that even MonoPerfCap gets almost right 3D capture results in the main camera view, the results in the reference view show the misalignments on the limbs of the human body.

Tab.\ref{tab:Comparison} shows the quantitative comparisons between our method and state-of-the-art methods using different evaluation metrics. We report the mean per joint position error (MPJPE), the Percentage of Correct Keypoints (PCK), and the Mean Intersection-Over-Union (mIoU) results. Benefiting from our multi-modal references, our method gets better performance than optimization-based methods and other data-driven methods.

\subsection{Evaluation} \label{paper:Evaluation}
We apply two ablation experiments. One verifies that it is more effective to apply the robust optimization stage, the other validates that the diligently-designed network structure and loss functions gain for the challenge human motions by comparing with other network structures.

\begin{figure}[tb]
	\centering
	\includegraphics[width=0.95\linewidth]{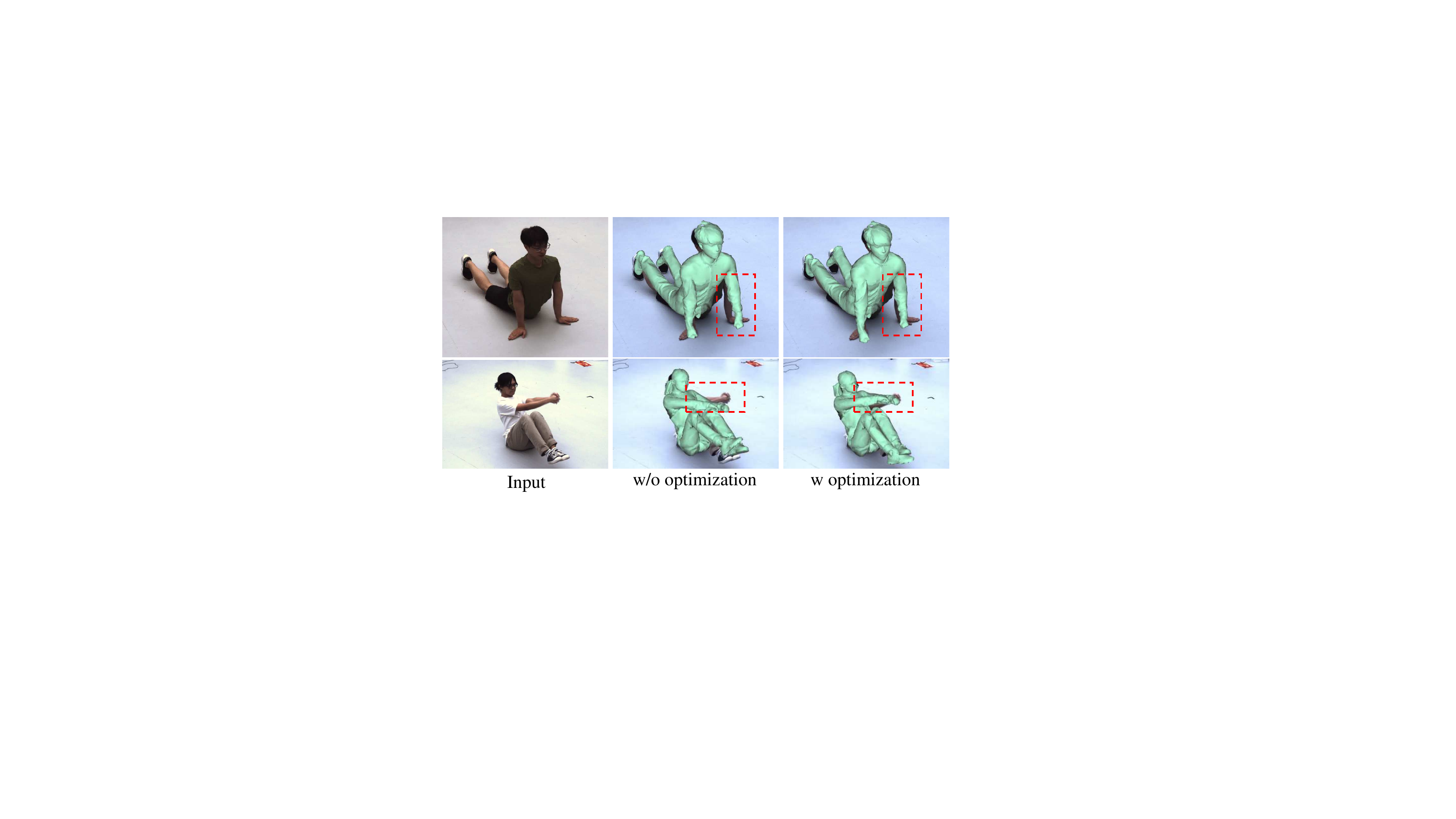}
	\caption{Evaluation for the optimization stage. The figure shows that our robust optimization stage improves the overlay performance.}
	\label{fig:fig_abla_optimize}
\end{figure}
\begin{table}[tb]
	\begin{center}
		\centering
		\caption{Quantitative evaluations on different optimization configurations.
		}
		\label{tab:optimization}
		\begin{tabular}{l|c|ccc}
			\hline
			Method         & MPJPE$\downarrow$         & PCK0.5$\uparrow$        & PCK0.3 $\uparrow$         \\
			\hline
			MonoPerfCap   & 134.7         & 77.4          & 65.6          \\
			Ours            & 106.5         & 92.8          & 82.2          \\
			Ours + optimization            & \textbf{52.6} & \textbf{96.6} & \textbf{87.4} \\
			\hline
		\end{tabular}
	\end{center}
\end{table}

\myparagraph{Evaluation on optimization.}
Tab.\ref{tab:optimization} shows the performance of models with or without the optimization module. The baseline 3D capture method is MonoPerfCap~\cite{MonoPerfCap}. The table demonstrates that whether the robust optimization stage is applied, our method outperforms MonoPerfCap. As illustrated in Fig.\ref{fig:fig_abla_optimize}, the obvious misalignment on the limb is corrected when the robust optimization stage is applied.

\begin{figure}[tb]
	\centering
	\includegraphics[width=0.95\linewidth]{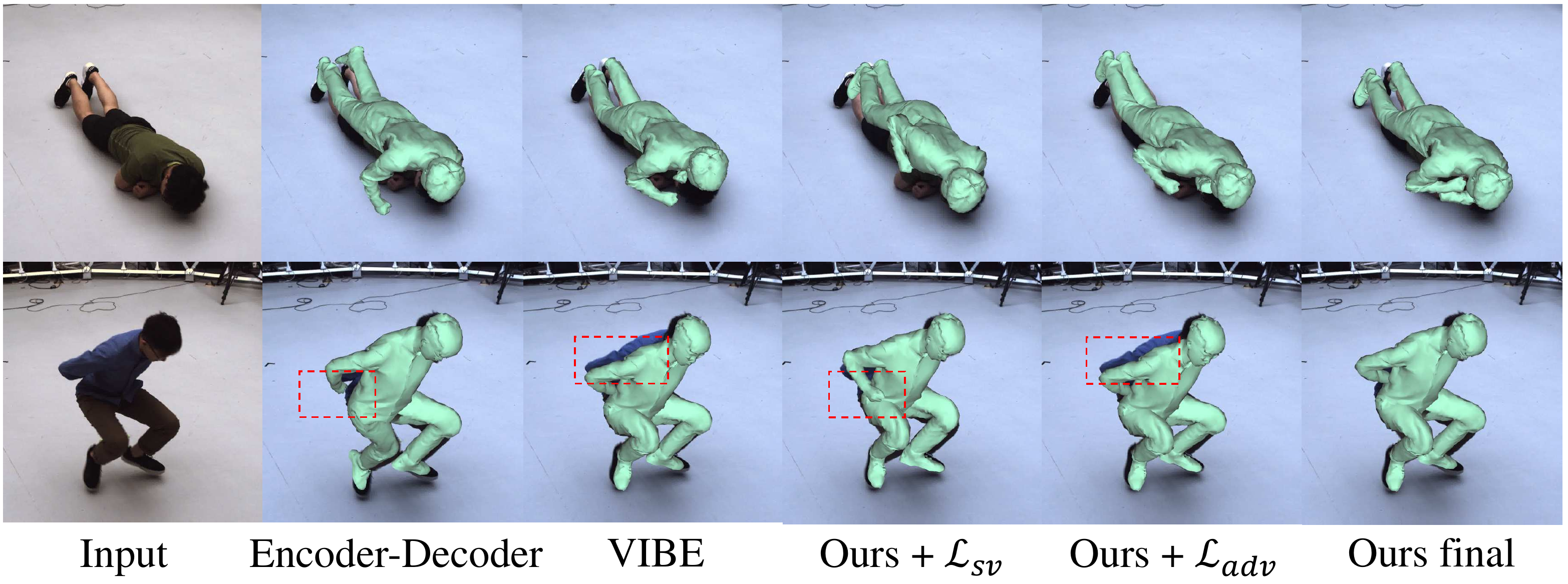}
	\caption{Evaluation of our network structure. The figure shows the effectiveness of both of our losses. Note that all experiments are applied with the robust optimization stage. Results of the full pipeline overlay more accurately with the input video frames.
	} 
	\label{fig:loss}
\end{figure}

\myparagraph{Evaluation on network structure.}
We experiment with several different network structures and loss design configurations, the results are demonstrated in Tab.\ref{tab:ablation_network}. The table shows that even only using sparse-view loss or adversarial loss, our method performs better than the simple encoder-decoder network structure without the attention pooling design. It also outperforms VIBE. When using both sparse-view loss and adversarial loss, our method gets  4\% to 5\% increase for PCK-0.5 and 12 to 32 decrease for MPJPE, compared with using only sparse-view or adversarial loss. The experiment illustrates multi-modal references contribute a lot to the improvement of results as illustrated in Fig. \ref{fig:loss}.

\begin{table}[t]
	\begin{center}
		\centering
		\caption{Quantitative evaluation of different network structure configurations. Our full pipeline achieves the lowest error.
		}
		\label{tab:ablation_network}
		\begin{tabular}{l|ccc}
			\hline
			Method                                               & MPJPE   $\downarrow$   & PCK0.5  $\uparrow$        & PCK0.3   $\uparrow$       \\
			\hline
			Encoder-Decoder                                       & 109.1         & 85.7          & 81.5          \\
			VIBE                                                  & 94.2          & 90.2          & 82.0          \\
			Ours  +  $\mathcal{L}_{sv}$                           & 64.3          & 92.5          & 84.1          \\
			Ours  +   $\mathcal{L}_{adv}$                         & 84.6          & 91.2          & 84.1          \\
			Ours  +  $\mathcal{L}_{sv}$ +   $\mathcal{L}_{adv}$   & \textbf{52.6} & \textbf{96.6} & \textbf{87.4} \\
			\hline
		\end{tabular}
	\end{center}
\end{table}

\section{Discussion}
\myparagraph{Limitation.}
As the first trial to explore challenging human motion capture with multi-modal references, the proposed ChallenCap still owns limitations as follows.
First, our method relies on a pre-scanned template and cannot handle topological changes like clothes removal.
Our method is also restricted to human reconstruction, without modeling human-object interactions.
It's interesting to model the human-object scenarios in a physically plausible way to capture more challenging and complicated motions.
Besides, our current pipeline turns to utilize the references in a two-stage manner.
It’s a promising direction to formulate the challenging motion capture problem in an end-to-end learning-based framework.

\myparagraph{Conclusion.}
We present a robust template-based approach to capture challenging 3D human motions using only a single RGB camera, which is in a novel two-stage learning-and-optimization framework to make full usage of multi-modal references.
Our hybrid motion inference learns the challenging motion details from various supervised references modalities, while our robust motion optimization further improves the tracking accuracy by extracting the reliable motion hints in the input image reference.
Our experimental results demonstrate the effectiveness and robustness of ChallenCap in capturing challenging human motions in various scenarios.
We believe that it is a significant step to enable robust 3D capture of challenging human motions,
with many potential applications in VR/AR, and performance evaluation for gymnastics, sports, and dancing.


{\small
\bibliographystyle{ieee_fullname}
\bibliography{egbib}
}

\end{document}